\documentclass[a4paper,twoside]{article}

\usepackage{epsfig}
\usepackage{subcaption}
\usepackage{calc}
\usepackage{amssymb}
\usepackage{amstext}
\usepackage{amsmath}
\usepackage{amsthm}
\usepackage{multicol}
\usepackage{pslatex}
\usepackage{apalike}
\usepackage{algorithm2e}
\usepackage[bottom]{footmisc}
\usepackage{SCITEPRESS}    
\usepackage{booktabs}
\usepackage{array}
\usepackage{multirow}
\usepackage{siunitx}
\usepackage{subcaption}
\usepackage{hyperref}

\newif\ifappendix
\appendixtrue

\begin{document}

\title{Reducing Bias in Pre-trained Models by Tuning while Penalizing Change}

\author{\authorname{Niklas Penzel\sup{1}\orcidAuthor{0000-0001-8002-4130}, Gideon Stein\sup{1}\orcidAuthor{0000-0002-2735-1842} and Joachim Denzler \sup{1}\orcidAuthor{0000-0002-3193-3300}}
\affiliation{\sup{1}Computer Vision Group, Friedrich Schiller University, Jena, Germany}
\email{\{niklas.penzel, gideon.stein, joachim.denzler\}@uni-jena.de}
}


\abstract{Deep models trained on large amounts of data often incorporate implicit biases present during training time.
If later such a bias is discovered during inference or deployment, it is often necessary to acquire new data and retrain the model.
This behavior is especially problematic in critical areas such as autonomous driving or medical decision-making.
In these scenarios, new data is often expensive and hard to come by.
In this work, we present a method based on change penalization that takes a pre-trained model and adapts the weights to mitigate a previously detected bias.
We achieve this by tuning a zero-initialized copy of a frozen pre-trained network.
Our method needs very few, in extreme cases only a single, examples that contradict the bias to increase performance.
Additionally, we propose an early stopping criterion to modify baselines and reduce overfitting.
We evaluate our approach on a well-known bias in skin lesion classification and three other datasets from the domain shift literature.
We find that our approach works especially well with very few images.
Simple fine-tuning combined with our early stopping also leads to performance benefits for a larger number of tuning samples.}

\onecolumn \maketitle \normalsize \setcounter{footnote}{0} \vfill

\section{\uppercase{Introduction}}
\label{sec:introduction}

There are many biases present in datasets used to train modern deep classifiers in part for sensitive applications, e.g., in skin lesion classification \cite{mishra2016overview}.
The models trained on such datasets learn to copy, i.e., reproduce these harmful biases.
This behavior leads to biased predictions on unseen data points and worse generalization.
In other words, applying such biased models in the real world leads tangible harm, e.g., racial biases in medical applications \cite{huang2022evaluation}.

\begin{figure}
  \centering
  \includegraphics[width=0.3\textwidth]{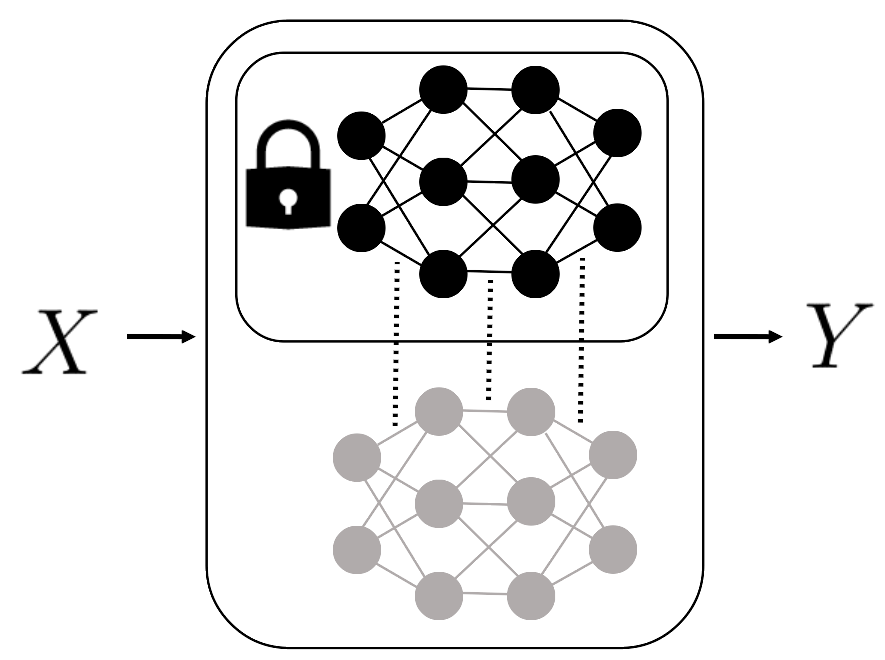}
  \caption{Architecture of the proposed method. We tune a zero-initialized change network (light grey) that is added to a frozen pre-trained model (black).}
  \label{fig:arch}
\end{figure}

In this work, we tackle the question of how we can reduce such harm by correcting biases in trained models.
Toward this goal, we propose a tuning scheme to reduce bias, relying on very few examples that specifically contradict the learned bias.
However, directly utilizing standard finetuning on such data would lead to overfitting, and performance would deteriorate.
Hence, we propose to penalize changes in the model weights harshly.
We achieve this by directly tuning a zero-initialized complementary change network with strong regularizations.
An overview of our approach can be seen in Figure~\ref{fig:arch}.

Additionally, we modify our baselines, and analyze whether an alternative early-stopping approach can help to improve performance.
We find that it is often enough to perform fine-tuning with small learning rates until the model corrects its behavior on the tuning data to increase performance on an unbiased test set.
This early stopping also falls in line with our motivation of minimum change.
Our work is based on the hypothesis that models trained on a biased training distribution still learn features related to the actual problem, which are sometimes overshadowed by biases found in the training data.
Reimers et al. give some evidence for this hypothesis in \cite{reimers2021conditional}.
They find that skin lesion classifiers learn biases as well as medically relevant features.

We use their work as a starting point and test our approach first on a well-known bias from the skin lesion classification domain.
Additionally, we evaluate our methods on multiple other debiasing and domain shift datasets contained in \cite{koh2021wilds}.
We find that penalizing change generally leads to improved performance on a less biased test set, especially if only single samples are used during the debiasing process.
Additionally, we find that our early stopping approach combined with baseline methods also leads to improvements for larger numbers of tuning samples and prevents overfitting in our evaluations.

\section{\uppercase{Related Work}}
\label{sec:related-work}

Many previous works propose debiasing approaches during train time, e.g., \cite{rieger2020interpretations,reimers2021conditional,tartaglione2021end}.
In \cite{reimers2021conditional}, Reimers et al. propose a secondary debiasing loss term $\mathcal{L}_{db}$ during training to penalize any conditional dependence between the learned representation and a bias term given the labels.
They show that such a conditional approach is beneficial compared to previous unconditional approaches.

Rieger et al. propose a similar secondary loss term in \cite{rieger2020interpretations}.
However, they penalize explanation errors instead.
If the model learns the right decisions but produces wrong explanations for its behavior, the loss term forces the model to correct the explanations by not relying on biases.
Roh et al. present FairBatch in \cite{roh2021fairbatch}, a sampling method that lowers biases during training by adaptively changing the data point sampling during stochastic gradient descent (SGD), formalizing SGD as a bilevel optimization problem. 
However, this approach needs access to the training data, unlike our strategy of penalizing change.
Roh et al. note that FairBatch can also be used to increase the fairness of pre-trained models simply by finetuning.

Furthermore, Tartaglione et al. \cite{tartaglione2021end} propose to use a regularization technique during training to prohibit the model from focusing on biased features.
They do this by introducing an information bottleneck where they employ a term entangling patterns corresponding to the same target class and a second term disentangling patterns that correspond to the same bias classes.

In contrast, our approach does not rely on changes in the training process but can instead be used to remove or lessen bias from pre-trained models directly.
This application is especially useful in the case that a new bias is later detected.
Related to this post hoc approach for debiasing is the work of Gira et al. \cite{gira2022debiasing}, where they show that it is possible to reduce biases in pre-trained language models by finetuning on debiased data.
They mitigate catastrophic forgetting by freezing the original model and adding only less than 1\% of the original parameters.
This approach is conceptionally related to our idea of minimal change.
However, we do not need to add additional parameters during the tuning process. 
Instead, we penalize the change itself.
Savani et al. \cite{savani2020intra} propose three methods of a class of debiasing they call intra-processing. 
Our method is also part of this debiasing class because we need access to the pre-trained model weights.
They propose a random perturbation, a layerwise optimization, and an adversarial debiasing approach.
However, in contrast to our approach, they directly try to optimize very specific fairness criteria.

Our definition of bias is closely related to domain shifts.
Hence, related to our work is the field of source-free domain adaptation (SFDA) \cite{yu2023comprehensive}.
Similarly to SFDA, we also do not need access to the source domain, only target, in our case less biased, domain.
However, SFDA specifically focuses on adaptation without target domain labels.
Hence, SFDA approaches often rely on self-training, e.g., \cite{liang2020we,qu2022bmd,lee2023feature}, or constructing a virtual source domain, e.g., \cite{tian2021vdm,ding2022proxymix}.
In contrast, we rely on the labels of the tuning data and consider only scenarios with very few examples.
We refer the reader to \cite{yu2023comprehensive} for more information about SFDA.

Our work is also related to the area of transfer learning, and finetuning \cite{perkins1992transfer,zhuang2020comprehensive}.
Here we want to mention the approach of side-tuning \cite{zhang2020side}.
Zhang et al. continuously adapt networks using additive side networks.
These networks can be useful when encoding priors or learning multiple tasks with the same backbone architecture.
They combine the side network outputs with the pre-trained output using alpha blending.
In contrast, our approach considers changes for all individual trainable weights.
Hence, even if our implementation via a change network is similar, we combine the weights in each layer respectively leading to an intertwined computational graph.
We empirically compare our approach to side-tuning in our experiments.

Also related are works that similarly to ours introduce regularization with respect to a given set of weights.
In \cite{chelba2006adaptation} and \cite{xuhong2018explicit} the authors respectively introduce distance-based regularizations for transfer learning.
Both works focus on $\ell_2$ regularization with respect to the pre-trained weights.
Xuhong et al. additionally investigate the $\ell_1$ norm.
In contrast, we also investigate a combination of both norms and find improvements compared to the individual variants.
In \cite{gouk2021distance} the authors utilize the maximum absolute row sum (MARS) norm $||.||_\infty$ and additionally introduce hyperparameters corresponding to the maximum allowable distances to the pre-trained weights.
They apply their approach to transfer learning tasks, while we specifically focus on reducing bias in a very few tuning samples scenarios.
Furthermore in \cite{barone2017regularization} the authors introduce tuneout for machine translation tasks, where the training is regularized by randomly exchanging weights in the network against the pre-trained versions during tuning similar to the popular dropout.

Another work related to ours and originating from the area of continual learning is memory aware synapses (MAS) \cite{aljundi2018memory}.
MAS estimates importance weights for each parameter in the network given the training data.
They then regularize changes in parameters correspondingly.
Hence, for new continual tasks, parameters that encode little information about previous tasks are tuned more.
In contrast, we want to change also the parameters that encode the bias.
Therefore, we penalize change in general.
Similarly to side-tuning, we empirically evaluate MAS in our described scenario and compare it against our approach.

\section{\uppercase{Method}}
\label{sec:method}

This section describes our approach and what we mean by penalizing change. 
We first describe how we define bias before deriving our updated loss function.

To start, let $f$ be a model parameterized by some pre-trained parameters $\theta$.
Given some problem space $\mathcal{X} \times \mathcal{Y}$, with inputs $x \in \mathcal{X}$ and labels $y \in \mathcal{Y}$, we assume that $f_\theta$ was trained on a biased sample from this space, i.e., on $(X, Y) \subset \mathcal{X}\times\mathcal{Y}$.
In other words, we do not see the whole distribution during train time.

However, we assume that $f_\theta$ works reasonably well on the correctly distributed test data $(\hat{X},\hat{Y}) \subseteq \mathcal{X}\times\mathcal{Y}$.
To be specific, we assume that our model $f_\theta$ is applied to data that correctly represents the latent distribution.

Given these data samples, we are interested in examples $(x,y) \in (\hat{X},\hat{Y})$ that are wrongly classified by $f_\theta$ presumably because they are part of the distribution not covered by $(X, Y)$.
In other words, we are interested in examples where
\begin{equation}
  f_\theta(x) \neq y,
\end{equation}
applies.
More specifically, we are interested in the necessary changes to the parameters $\theta$ that correct the mistake, i.e.,
\begin{equation}
\label{eq:correction}
  f_{\theta + \theta'}(x) = y.
\end{equation}
For a given problem, it is possible that multiple such $\theta'$s exist.
This is especially true given the non-convex loss surfaces of neural networks.

However, we are interested in the specific changes $\theta'$ that are minimal in some sense, i.e., that change the original parameters $\theta$ the least, to preserve pre-trained knowledge.
This change could either be minimal with respect to some norm, e.g., $||\theta'||_2$, or the number of parameters changed.

Without loss of generality with respect to the norm used, we are interested in the loss function $\mathcal{L}_{mc}$ composed of two terms:
First, the original loss function $\mathcal{L}$ is used to train the original set of parameters $\theta$, and second, the minimization constraint for some norm of the parameter change $||.||$.
We define $\mathcal{L}_{mc}$ as
\begin{equation}\label{eq:loss}
  \mathcal{L}_{mc} = \mathcal{L}(f_{\theta + \theta'}(x), y) + \lambda ||\theta'||,
\end{equation}
where $\lambda$ is a hyperparameter similar to the standard weight decay parameter that describes how strongly we constrain the change in parameters.

To optimize $\mathcal{L}_{mc}$, we are using the gradient with respect to $\theta'$
\begin{equation}\label{eq:grad}
  \nabla_{\theta'} \mathcal{L}_{mc} = \frac{\partial}{\partial \theta'} \mathcal{L}(f_{\theta + \theta'}(x), y) + \lambda \frac{\partial}{\partial \theta'} ||\theta'||.
\end{equation}

Equations \eqref{eq:loss} and \eqref{eq:grad} can easily be adapted to multiple wrongly classified examples, i.e., batches of data, by utilizing the standard notation for stochastic gradient descent.

Our derivation of the objective function $\mathcal{L}_{mc}$ is similar to the classical formalization of parameter norm penalties. 
See, for example, Section 7.1 in \cite{goodfellow2016deep}.
However, we penalize the parameter change for some fixed $\theta$.
Intuitively, we detach the original model from the computational graph for automatic differentiation and instead add a zero-initialized change network of the same architecture.
On this change model, our statements are equivalent to standard parameter norm penalization.

In our experiments, we use $\ell_1$ and $\ell_2$ norm or combinations thereof.
These two norms realize two different notions of ``small'' change.
The $\ell_2$ norm leads to changes with small Euclidean norm, while $\ell_1$ norm can lead to sparse solutions, i.e., change in fewer parameters \cite{goodfellow2016deep}.

\subsection{\uppercase{Implementation Details and Stopping Criterion}}
\label{sec:impl-details}
Following the observation that we can model our approach non-destructively as a change network, we give some implementational details in this section and propose a stopping criterion as additional regularization.
For reference, we use the framework PyTorch \cite{paszke2019pytorch}.

The part of Equation~\eqref{eq:loss} that is interesting during implementation is the parameter sum in $f_{\theta + \theta'}$.
To simplify the implementation of this parameter sum for many neural network architectures, we utilize the following observation.

Let $g_1, g_2$ be two linear transformations given by 
\begin{equation}
  g_i(x) = W_i \cdot x + b_i,
\end{equation}
with parameters $W_i$ and $b_i$.
Then, $g_1(x) + g_2(x) = (W_1 + W_2) \cdot x + (b_1 + b_2)$ applies.
This observation holds equivalently for the convolution operation used in many architectures because it is a linear transformation.
Furthermore, batch normalization \cite{ioffe2015batch} is linear with respect to the scaling parameters.
Hence, this observation holds also for batch normalization layers.

Using this decoupling of the sum of weights, we can calculate the necessary output of a combined layer by calculating the output of both the pre-trained and the change layers separately.
However, this only applies to layers that perform a linear transformation.
Hence, we have to add the outputs together before propagating them through the non-linearity to the next layer.
Nevertheless, this enables us to circumvent error-prone editing of the computational graph by simply performing a layerwise output sum.
The proposed implementation is related to the idea of side networks in side-tuning \cite{zhang2020side}.
However, while side networks are combined on the output level of the whole network, we introduce the learned changes in the layer-wise fashion described above.
Our zero-initialization of $\theta'$ is related to how ControlNet models are attached to the pre-trained networks in \cite{zhang2023adding}.

Another practical consideration is the duration of the tuning process. 
In other words, how many update steps are necessary to correct the bias but prohibit overfitting?
To tackle this problem, we propose an early stopping regime \cite{yao2007early} inspired by our problem motivation (see Equation~\eqref{eq:correction}).
Our heuristic is to generally stop the training once the network correctly predicts the examples in the tuning batch.
This is different from standard early stopping approaches in two ways: First, the stopping metric is not the loss function, and second, we evaluate the criterion on the tuning data itself.

Early stopping is another form of regularization to reduce overfitting \cite{goodfellow2016deep}.
However, note that given the zero-initialization of our described change network, our penalization of the change in the first update step is zero.
If the batch of data we use to tune the network is small, it can happen that the first step would be enough to correct the network behavior for all tuning samples.
Hence, stopping when the network corrects the predictions could remove the influence of our change penalization.
Therefore, we introduce a parameter $\epsilon$, which determines the minimum amount of steps to train after the model correctly classifies the tuning data.
In other words, $\epsilon > 0$ ensures that even if the model overshoots after the first update step, the change can be penalized correctly in the successive steps leading to small changes in the parameters.
This early stopping scheme enables us to extend arbitrary baselines and reduce overfitting.
Hence, we also report updated fine-tuning, side-tuning \cite{zhang2020side} and MAS \cite{aljundi2018memory} in our experiments.

\section{\uppercase{Experiments}}
\label{sec:experiments}

We evaluate our approach empirically on different problems.
Toward this goal, we selected different data sources that fit our definition of bias defined above.
In other words, for each of the following classification problems, we can clearly specify a difference between the training and testing distribution with a measurable decrease in network performance.
However, first, we detail our general experiment setup and hyperparameter settings before describing the selected datasets and results.

\subsection{\uppercase{General Setup}}
\label{sec:setup}
Our general experiment setup is as follows:
First, we draw a batch of $b$ samples from our set of bias-contradicting images, i.e., images that are part of the test distribution and wrongly predicted by the network.
Second, we tune the model using our proposed change penalization, fine-tuning, side-tuning \cite{zhang2020side}, or MAS \cite{aljundi2018memory}.
We adapt our baselines using our early stopping scheme and setting $\epsilon$ to zero.
See Appendix~\ref{app:overfit} for larger $\epsilon$ values simulating standard tuning.

In all our experiments, we perform five runs per trained model using different non-overlapping batches of tuning data.
We investigate number of tuning samples $b$ between 1 and 32.
Furthermore, for each dataset, we repeat this tuning process for three different models trained on the corresponding biased training distribution and report results averaged over both these sources of randomness.
Where applicable, we also redraw the train-val-test split before initially training our models.
In our experiments, ImageNet \cite{russakovsky2015imagenet} pre-trained ResNet18 \cite{he2015deep} models are used to train the initially biased models.
In our evaluation, we report the difference in balanced accuracy on the unbiased test set after the tuning process if not stated differently.
Here balanced accuracy means the average of the accuracy scores calculated per class, e.g., \cite{brodersen2010balanced}, and we use it to ensure an accurate estimation of model performance in the presence of dataset imbalances
Given that we report the difference before and after the tuning, values larger than zero indicate an increase in performance, while values below zero represent a decrease.
In all our figures, we report the mean and standard deviation.

For the initial biased training process, we train the models using SGD with a learning rate $\gamma$ set to 1e-3, a momentum of $0.9$, and standard $\ell_2$ weight decay of 5e-4.
As mentioned above, we compare our approach to three other approaches adapted to our problem setting.
For side-tuning, we follow the best setup described in \cite{zhang2020side}, and initialize the side-network with as the pre-trained network.
Similarly, for MAS \cite{aljundi2018memory}, we follow the author's suggestion and use their $\ell_2$ approximation of the importance weights $\Omega$ and set the corresponding regularization parameter to 1.
For standard fine-tuning and also for the other comparison methods, we use a reduced learning rate of 1e-5 together with the otherwise unchanged parameters of the pre-training.

However, we adapt all three baselines to use $\epsilon = 0$ to stop the tuning process early.
Note that this adaption is already a form of change regularization and, therefore, encapsulates some of our intuition about minimum change.
We apply this early stopping for two main reasons:
Firstly, it ensures that the methods fit closely to the problem we are trying to solve.
Second, without this adaptation, classical fine-tuning is prone to overfit on the 1 to 32 tuning examples leading to a degradation in performance.
We show this behavior in Appendix~\ref{app:overfit}.

For our approach, we report results for the following hyperparameter setting:
First, we select a combination of the $\ell_1$ and $\ell_2$ norms as our change penalization norm of choice to encode the two intuitions of small change discussed in Section~\ref{sec:method}.
Specifically, we use $0.5 \cdot (\ell_1 + \ell_2)$.
Additionally, we set $\lambda$ to 1.0 and use $\epsilon = 0$ to ensure comparability with our selected baselines.
Appendix~\ref{app:hp-selection} includes a limited ablation study of different parameter settings for our approach and the melanoma classification task described in Section~\ref{sec:mela}.

\subsection{\uppercase{ISIC archive: Melanoma}}
\label{sec:mela}

\begin{figure*}[t]
    \centering
    \includegraphics[width=\textwidth, trim=2cm 0cm 2cm 1cm, clip]{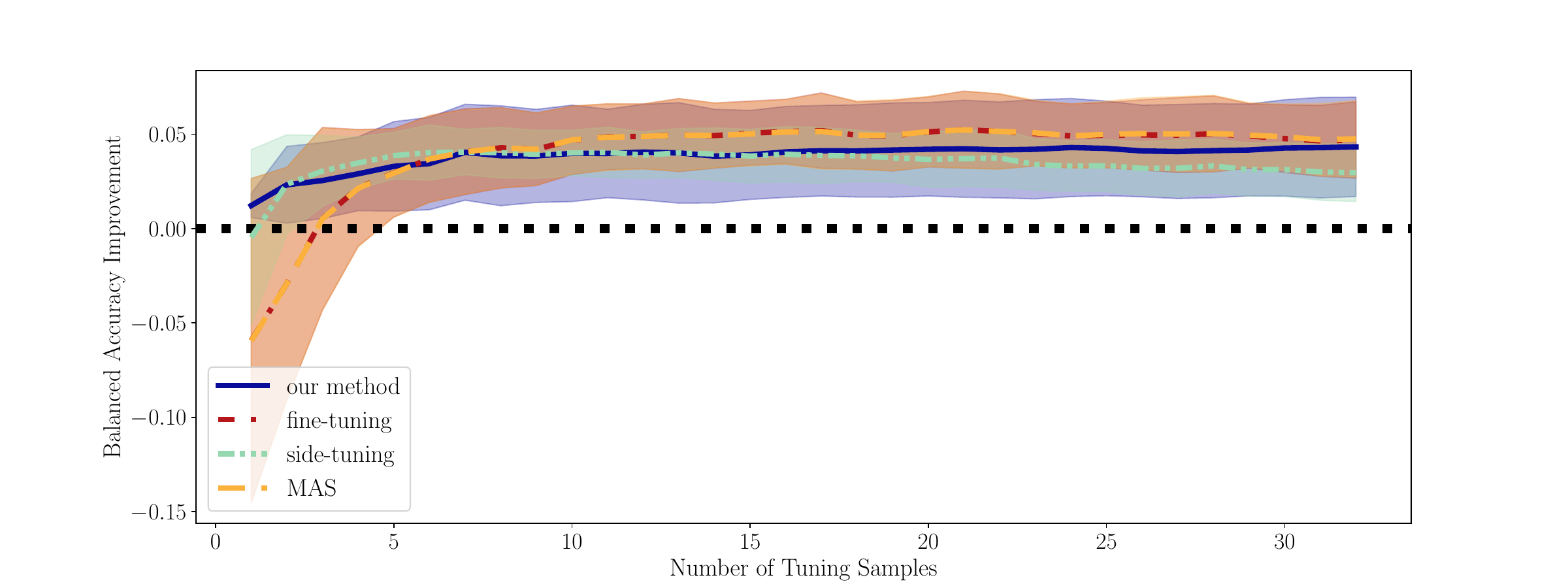}
    \caption{
    The difference in accuracy after debiasing using our approach versus three baseline methods combined with our stopping scheme on the melanoma classification task. 
    }
    \label{fig:isic-results}
\end{figure*}

\begin{figure*}[t]
    \centering
    \includegraphics[width=\textwidth, trim=2cm 0cm 2cm 1cm, clip]{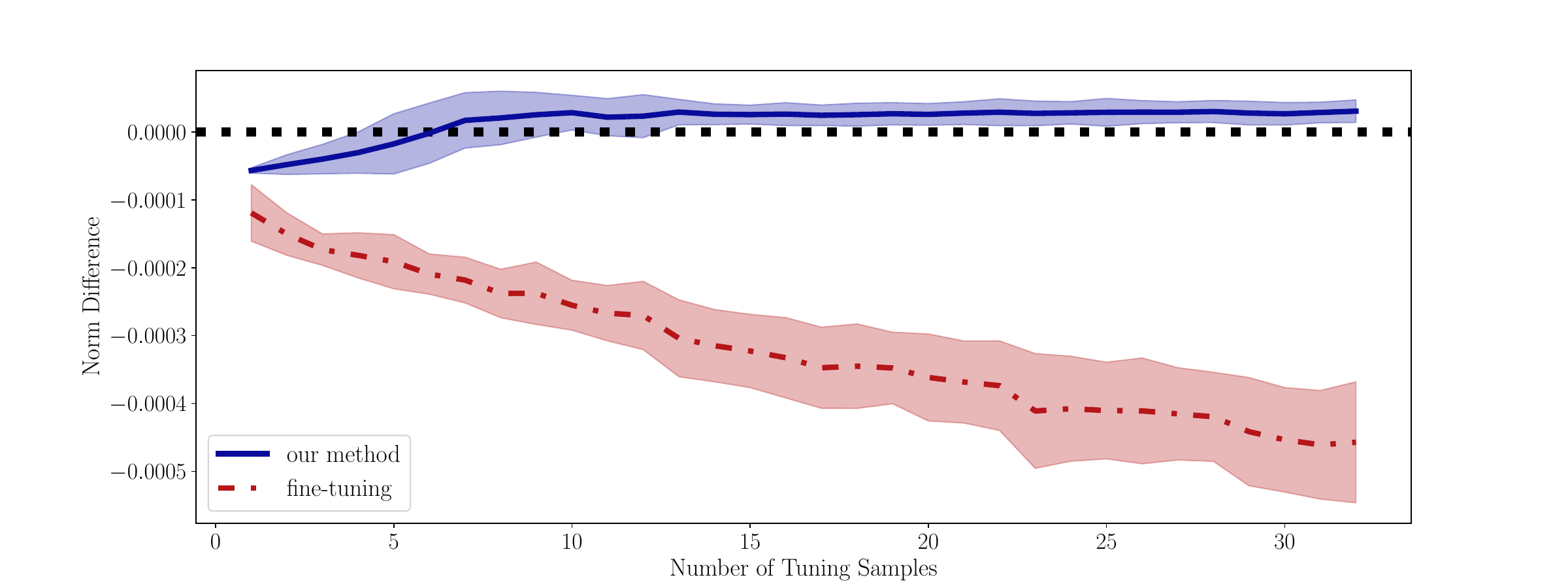}
    \caption{
    Difference of the Euclidean norm of the model parameter vectors before and after the debiasing. 
    Note that side-tuning \cite{zhang2020side} and MAS \cite{aljundi2018memory} lead to larger changes and overshadow the difference between our approach and fine-tuning. Hence, they are omitted here.
    }
    \label{fig:isic-norms}
\end{figure*}

\paragraph{Dataset:}
The ISIC archive \cite{isic-archive}, a public collection of skin lesion data from various sources, includes the SONIC dataset \cite{scope2016study}.
In this study, Scope et al. analyzed benign nevi in children.
However, they used large colorful patches to cover neighboring skin lesions, which are often included in the images.
This practice results in a colorful bias in skin lesion image classification.
All $\sim$10K images of the original study that contain colorful patches are of the same diagnosis - benign nevi.
We utilize this fact to construct a simple binary dataset for melanoma classification from the ISIC archive containing the diagnoses of melanoma and benign nevi.
Some of the selected nevi images contain the described patches.
Additionally, we balance the two classes, melanoma and benign nevi, to focus on the colorful patches bias.
We use this dataset to train our initially biased models.
On a test set following the training distribution, the model achieves an average accuracy of 0.865.

We then construct an unbiased distribution by randomly pasting previously extracted colorful patches on images of the melanoma class.
To achieve this, we use patch segmentations created by Rieger et al. \cite{rieger2020interpretations}.
Figure~\ref{fig:isic-examples} in Appendix~\ref{app:isic-examples} shows some examples from our biased training and less biased testing distribution.
On the latter distribution, our biased models drop to an average accuracy of $0.801$, i.e., a decrease of 6.4 points.
We then use the setup described in Section~\ref{sec:setup} together with images containing melanomata and colorful patches to perform debiasing.

\paragraph{Results}

Figure~\ref{fig:isic-results} visualizes the results of our debiasing experiments.
We can see that for very small numbers of tuning samples, our approach of parameter change penalization is beneficial and leads to an improvement in performance on the less biased test distribution given as few as a single image.
However, for larger batch sizes, the baselines combined with our stopping scheme reach similar and even slightly better results on average.

To investigate the similar performance further, we additionally visualize $||\theta||_2 - ||\theta + \theta'||_2$, i.e., the difference in the Euclidean norm of the model parameter vectors before and after the debiasing for our approach and fine-tuning.
Figure~\ref{fig:isic-norms} shows that even though both methods achieve a similar increase in performance for larger numbers of tuning samples, they strive towards different optima.
In fact, using our approach, the Euclidean norm of the parameter vector increases, which is possible because we penalize change and do not perform classical weight decay.
We can also observe that our approach leads to, on average, smaller changes in the parameters $\theta$ due to our regularization.

\subsection{\uppercase{CelebA: Haircolor}}
\label{sec:celebA}

\begin{figure*}
    \centering
    \includegraphics[width=\textwidth, trim=2cm 0cm 2cm 1cm, clip]{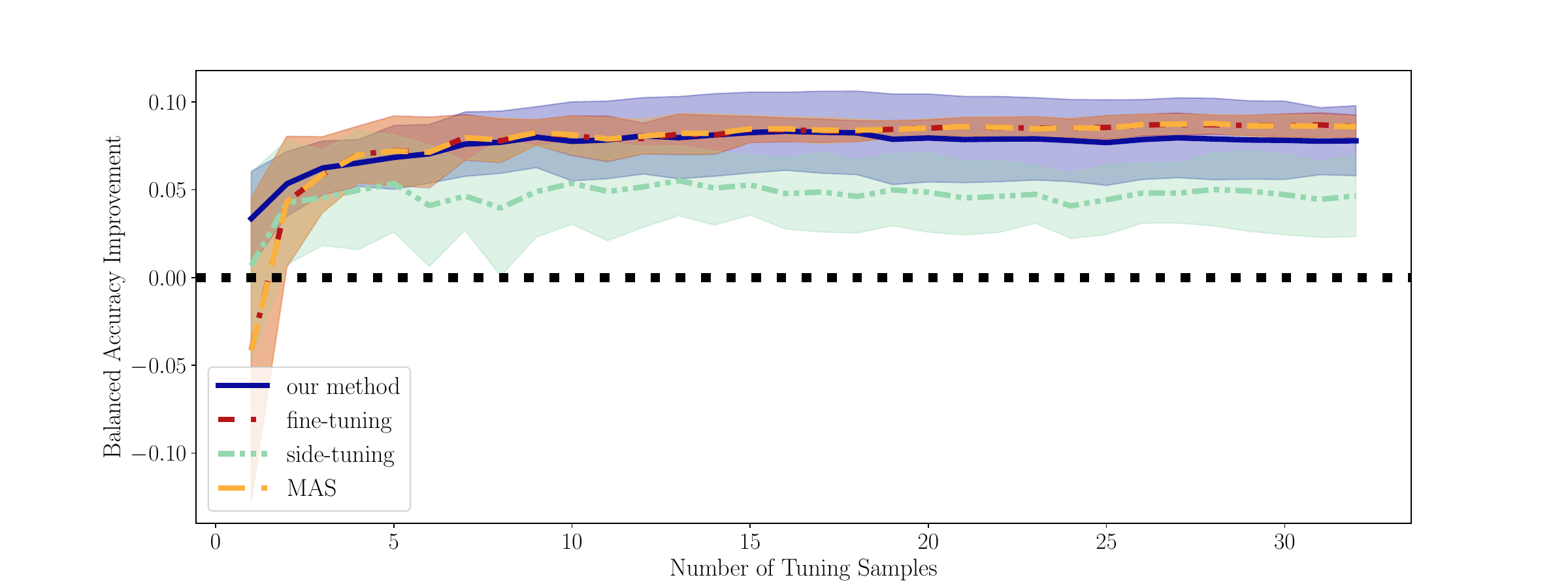}
    \caption{
    The difference in accuracy after debiasing using our approach versus three baseline methods combined with our early stopping scheme on the celebA \cite{liu2015faceattributes} classification task. 
    }
    \label{fig:celebA-results}
\end{figure*}

\paragraph{Dataset}

The second dataset we chose is the celebA dataset \cite{liu2015faceattributes} consisting of over 200K images picturing the faces of celebrities and annotated with facial attributes.
We use the implementation from \cite{koh2021wilds} and construct a simple binary classification problem where the task is to decide whether the hair color in an image is blond.
However, in our training distribution, all people with blond hair are annotated as female in celebA, while we balance not blond people between male and female celebrities.
Our biased models achieve, on average, a balanced accuracy of  0.947 on this training distribution.
This performance drops to 0.779 on our test distribution, where no correlation between hair color and annotated sex exists.
For debiasing, we only use images of celebrities labeled as male with blond hair color.
Appendix~\ref{app:celebA} contains additional information and example images of our data distributions. 

\paragraph{Results}

The second task posed by the described celebA \cite{liu2015faceattributes} subset is conceptionally similar to the first problem we analyzed.
Again we investigate a one-sided binary bias.
Figure~\ref{fig:celebA-results} visualizes our results.
Similar to our observations in Section~\ref{sec:mela}, here, our approach of change penalization leads to improvements as well, especially for smaller numbers of tuning samples.
Given larger batch sizes, the change regularization implemented by our stopping criterion combined with the baseline methods also results in a performance increase.
MAS and fine-tuning perform near identical and lead, on average, to the highest increases for a number of 32 tuning samples.
In contrast, side-tuning is outperformed by the other approaches.
Nevertheless, it still leads to an increase in balanced accuracy. 

\subsection{\uppercase{Waterbirds: Bird Type}}
\label{sec:waterbirds}

\begin{figure*}[t]
    \centering
    \begin{subfigure}{\textwidth}
    \centering
        \includegraphics[width=\textwidth, trim=2cm 0cm 2cm 1cm, clip]{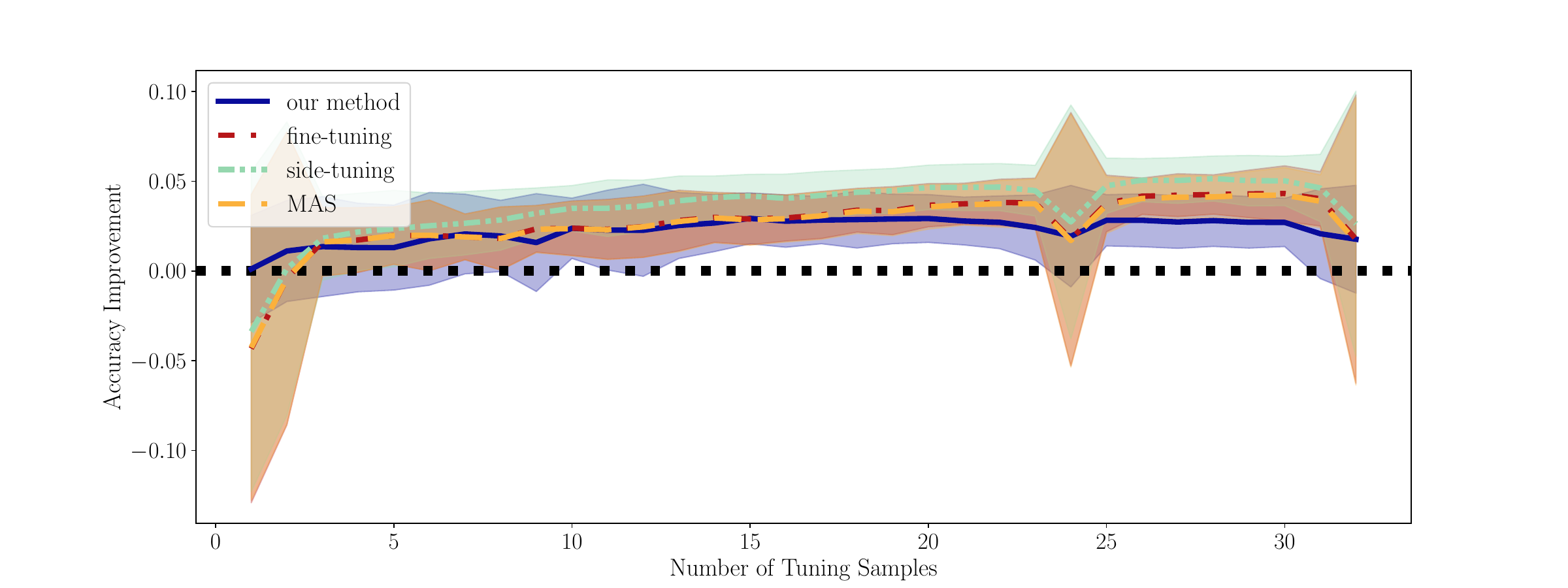}
        \caption{Difference in accuracy.}
        \label{fig:waterbirds-acc}
    \end{subfigure}
    \begin{subfigure}{\textwidth}
    \centering
        \includegraphics[width=\textwidth, trim=2cm 0cm 2cm 1cm, clip]{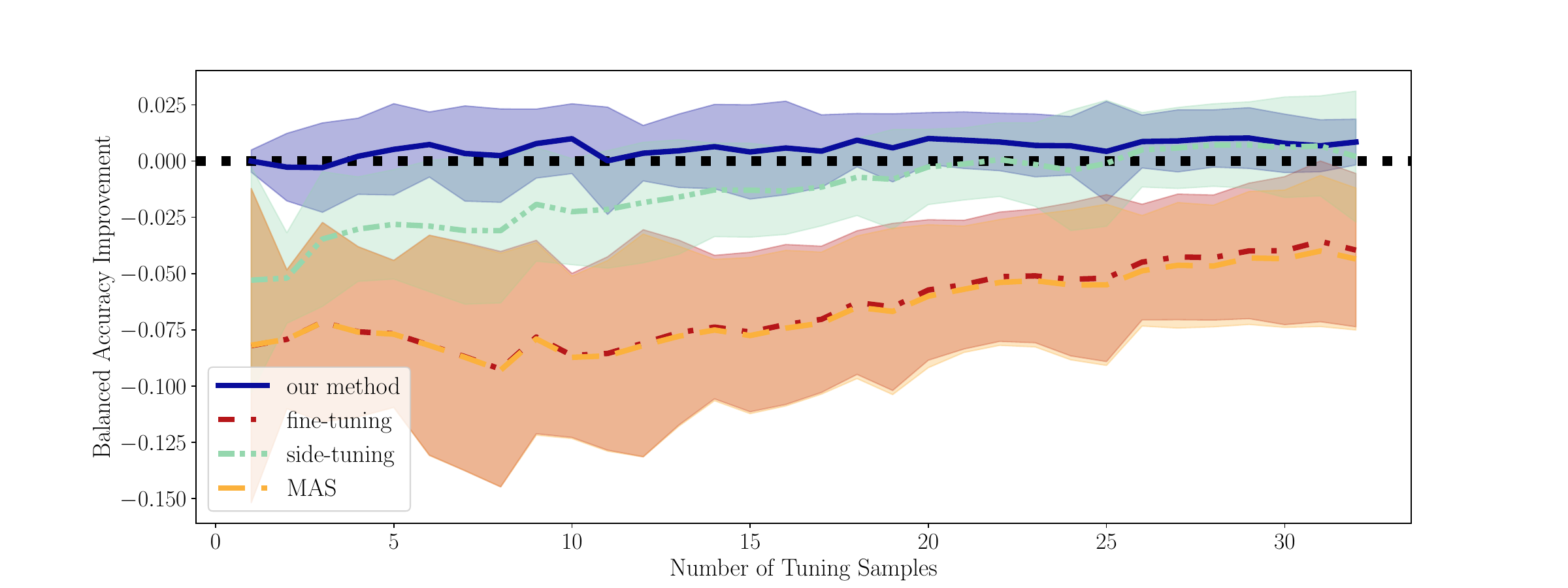}
        \caption{Difference in balanced accuracy.}
        \label{fig:waterbirds-bacc}
    \end{subfigure}
    \caption{The differences in accuracy and balanced accuracy after debiasing using our approach versus three baseline methods together with our early stopping scheme on the waterbirds dataset \cite{sagawa2019distributionally}. 
    }
    \label{fig:waterbirds-results}
\end{figure*}

\paragraph{Dataset}

Following the observation for the last two datasets, i.e., that heavy change penalization can help to improve simple one-sided binary biases, we now investigate a two-sided bias.
Toward this goal, we use the waterbirds dataset \cite{sagawa2019distributionally} with the implementation from \cite{koh2021wilds}.
This dataset is constructed by combining CUB200 \cite{wah2011caltech} birds together with Places \cite{zhou2017places} backgrounds.
The task is to classify whether a displayed bird is a land or water-based species.
However, in the training distribution, there is a high correlation between the label and a corresponding background containing either land or water.
Note that this means, in contrast to the previous datasets we analyzed, that there is still an overlap, i.e., some land birds are combined with water images.
Nevertheless, in the unbiased test distribution, this correlation vanishes.
The balanced accuracies of our biased models, therefore, decrease from 0.931 to 0.786 on average.
See Appendix~\ref{app:waterbirds} for example images from this dataset.
Note that neither the training nor test data is class-wise balanced.
Hence, we compare the results for standard accuracy and balanced accuracy.
This comparison enables us 
Finally, for debiasing, we use images from the validation set, which follows the test distribution.

\paragraph{Results}

For the more complicated two-sided bias of the waterbirds dataset \cite{sagawa2019distributionally}, we analyze both the difference in accuracy and balanced accuracy.
Figure~\ref{fig:waterbirds-acc} shows that all methods lead to a similar increase in standard accuracy.
However, our approach seems to perform slightly worse for larger numbers of tuning samples.
Yet, Figure~\ref{fig:waterbirds-bacc} shows that our approach does not degrade the balanced accuracy.
In contrast, while fine-tuning and MAS lead to improvements in the unbalanced accuracy metric, they drop the balanced accuracy by as much as $0.1$.
Both methods perform worse for smaller amounts of tuning samples.
Side-tuning still drops the performance for nearly all numbers of tuning samples but performs in between our approach and the other two baselines.

\subsection{\uppercase{Camelyon17: Cancer Tissue}}
\label{sec:camelyon17}

\begin{figure*}
    \centering
    \includegraphics[width=\textwidth, trim=2cm 0cm 2cm 1cm, clip]{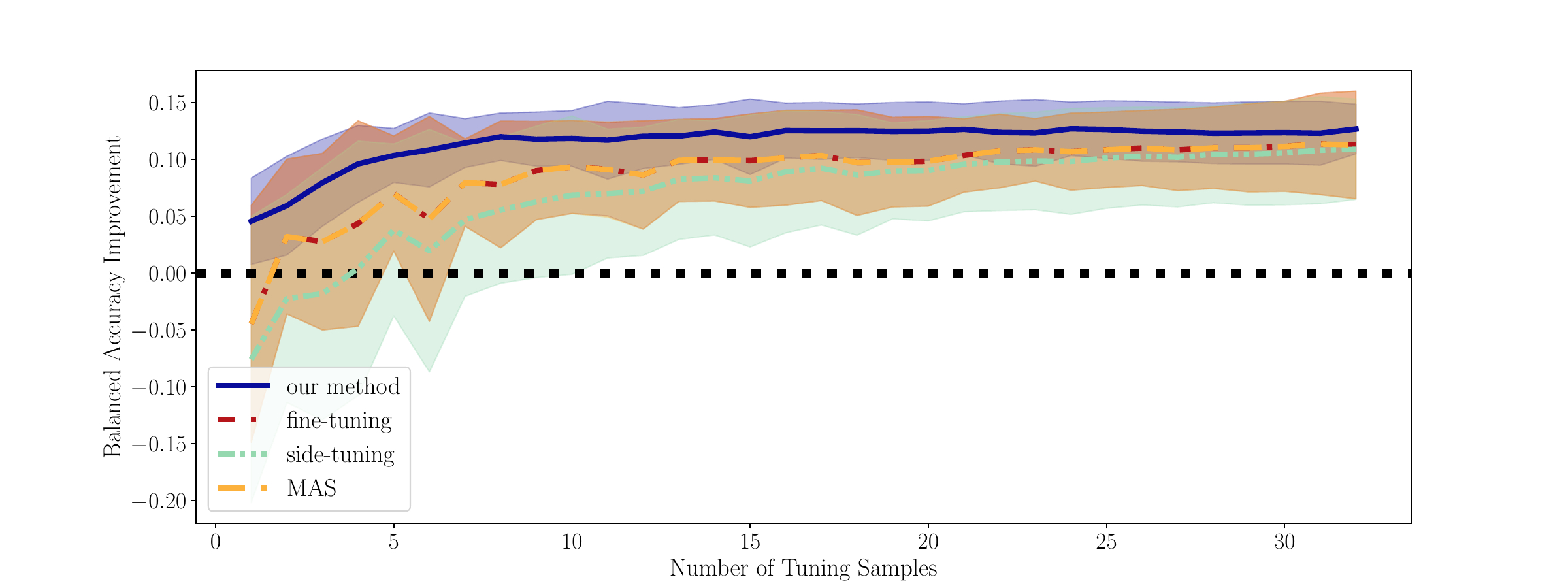}
    \caption{The difference in accuracy after debiasing using our approach versus three baseline methods together with our early stopping scheme on the camelyon17 dataset \cite{bandi2018detection}. 
    }
    \label{fig:camelyon17-results}
\end{figure*}

\paragraph{Dataset}
The last dataset of our analysis is the camelyon17 dataset \cite{bandi2018detection}.
We select this domain shift dataset given the similarity of our definition of bias and domain shift in general.
We use the implementation from \cite{koh2021wilds}, where the task is to classify whether images of tissue slides contain cancerous tissue in the image center.
However, while the training data is collected from four different hospital sites, the test data is captured in a disjunct institution leading to a clearly visible domain shift.
See Appendix~\ref{app:camelyon17} for examples.
Given this domain shift, the performances of our pre-trained models deteriorate from 0.996 on the train distribution down to 0.786 on the test distribution.
For debiasing, we use held-out images from the test split.

\paragraph{Results}

Figure~\ref{fig:camelyon17-results} visualizes the results for our last set of experiments.
Here our approach of change penalization performs best on average for all tested numbers of tuning samples.
Given only one tuning sample, we can increase performance by around 0.05.
In contrast, the baselines decrease performance for such little data.
However, for a larger number of tuning images, all methods increase the model performance on the less biased test distribution.
Again we observe that MAS and fine-tuning combined with our stopping scheme lead to similar results.

\section{\uppercase{Conclusions}}
\label{sec:conclusions}

In this work, we motivate the idea of debiasing pre-trained image classification models using heavy penalization of weight changes coupled with tuning data contradicting learned biases.
This approach follows the observation that models learn a combination of biases and meaningful features \cite{reimers2021conditional}.
To penalize change, we propose two methods of change regularization: one based on an additional loss function term and one early stopping scheme.
The latter of which can easily be added to existing transfer learning approaches to lessen overfitting.

Our approach of adding a regularization term leads to increased performance on less biased test distributions of varied classification tasks in our extensive empirical evaluation.
This observation holds true for as little as one image from the test sets.
Nevertheless, for larger numbers of tuning samples, we conclude that it is often enough to utilize simple fine-tuning together with our early stopping scheme, which leads to similar or better performance benefits.

A future research direction is other penalization metrics that could lead to further benefits, e.g., the spectral norm to force low-rank parameter changes.

\bibliographystyle{apalike}
{\small
\bibliography{sources}}

\ifappendix
    \section*{\uppercase{Appendix}}

    \subsection{\uppercase{Colorful Patches Examples}}
    \label{app:isic-examples}

    Figure~\ref{fig:isic-examples} displays examples of the original biased training data and the melanomata with inpainted colorful patches we added to the test distribution.
    The resulting less-biased test distribution contains benign nevi and melanomata images with colorful patches.

    \begin{figure}
    \centering
    \begin{subfigure}{0.45\textwidth}
    \includegraphics[width=0.24\textwidth]{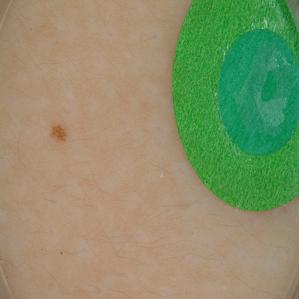}
    \includegraphics[width=0.24\textwidth]{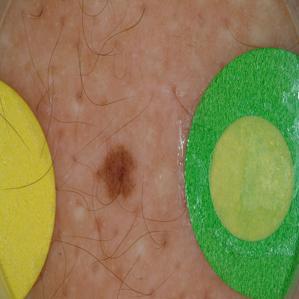}
    \includegraphics[width=0.24\textwidth]{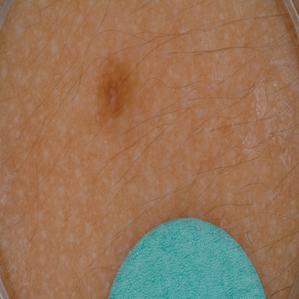}
    \includegraphics[width=0.24\textwidth]{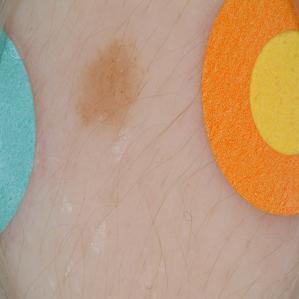}
    \caption{Original images of class nevus.}
    \end{subfigure}\\
    \begin{subfigure}{0.45\textwidth}
    \includegraphics[width=0.24\textwidth]{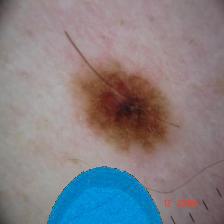}
    \includegraphics[width=0.24\textwidth]{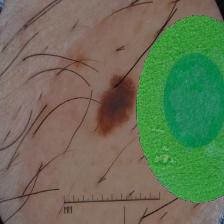}
    \includegraphics[width=0.24\textwidth]{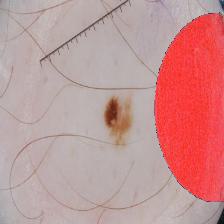}
    \includegraphics[width=0.24\textwidth]{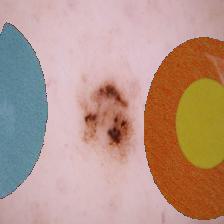}
    \caption{Artificially inpainted melanoma images.}
    \end{subfigure}
    \caption{Some examples from biased training and less biased test distribution.}
    \label{fig:isic-examples}
    \end{figure}

    \subsection{\uppercase{CelebA Examples}}
    \label{app:celebA}

    \begin{figure}
        \centering
        \includegraphics[width=0.48\textwidth]{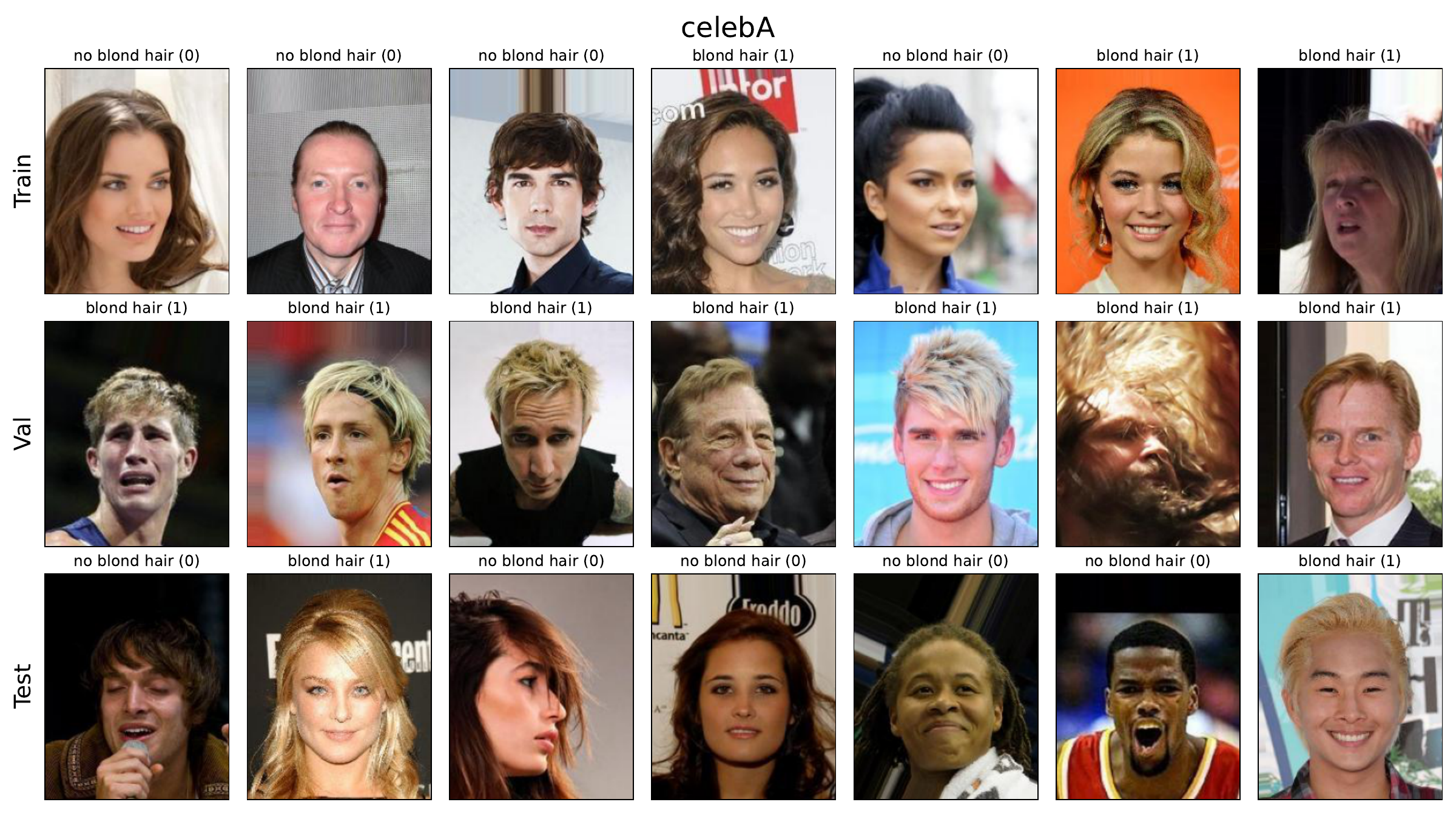}
        \caption{
        Examples contained in the celebA dataset \cite{liu2015faceattributes}. 
        All examples of class blond are female in the train split (first row).
        The row titled ``val'' contains tuning examples, i.e., men with blond hair.
        The final row contains the test distribution, where men and women are uniformly distributed in both classes.
        }
        \label{fig:celebA-examples}
    \end{figure}

    Figure~\ref{fig:celebA-examples} displays examples from the celebA dataset \cite{liu2015faceattributes} described in Section~\ref{sec:celebA}.
    We use the implementation provided by \cite{koh2021wilds} but rebalance and resplit the dataset.
    Hence, the resulting dataset is classwise balanced.
    
    \subsection{\uppercase{Waterbirds Examples}}
    \label{app:waterbirds}

    \begin{figure}
        \centering
        \includegraphics[width=0.48\textwidth,trim=2cm 2cm 0cm 3.5cm, clip]{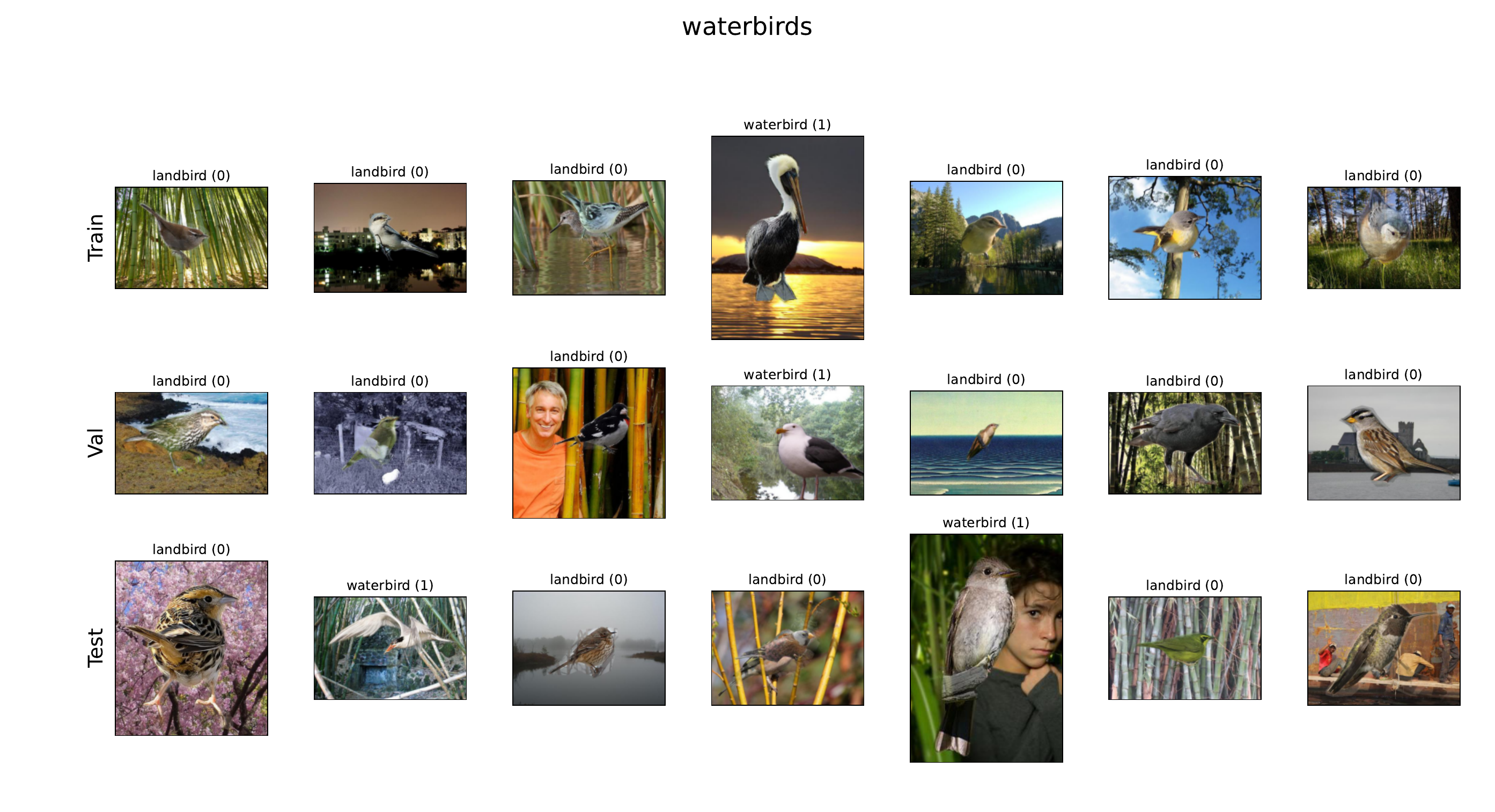}
        \caption{
        Examples contained in the waterbirds dataset \cite{sagawa2019distributionally}. 
        In the training distribution, birds typically found in water-based environments correlate more with water backgrounds, while land birds are often seen in front of land-based backgrounds.
        For the validation, i.e., our tuning split and the test distribution, this correlation is nearly zero.
        }
        \label{fig:waterbirds-examples}
    \end{figure}

    Figure~\ref{fig:waterbirds-examples} contains examples of the waterbirds dataset \cite{sagawa2019distributionally} described in Section~\ref{sec:waterbirds}.
    This dataset is constructed with birds from the CUB-200 dataset \cite{wah2011caltech} and backgrounds from the Places dataset \cite{zhou2017places}.
    In the training-split water and land-based birds correlate highly with corresponding backgrounds.
    In contrast, the test distribution is more balanced, and the backgrounds and birds are uncorrelated.
    We use the implementation provided by \cite{koh2021wilds}.

    \subsection{\uppercase{Camelyon17 Examples}}
    \label{app:camelyon17}

    \begin{figure}
        \centering
        \includegraphics[width=0.48\textwidth,trim=2cm 2cm 0cm 4cm, clip]{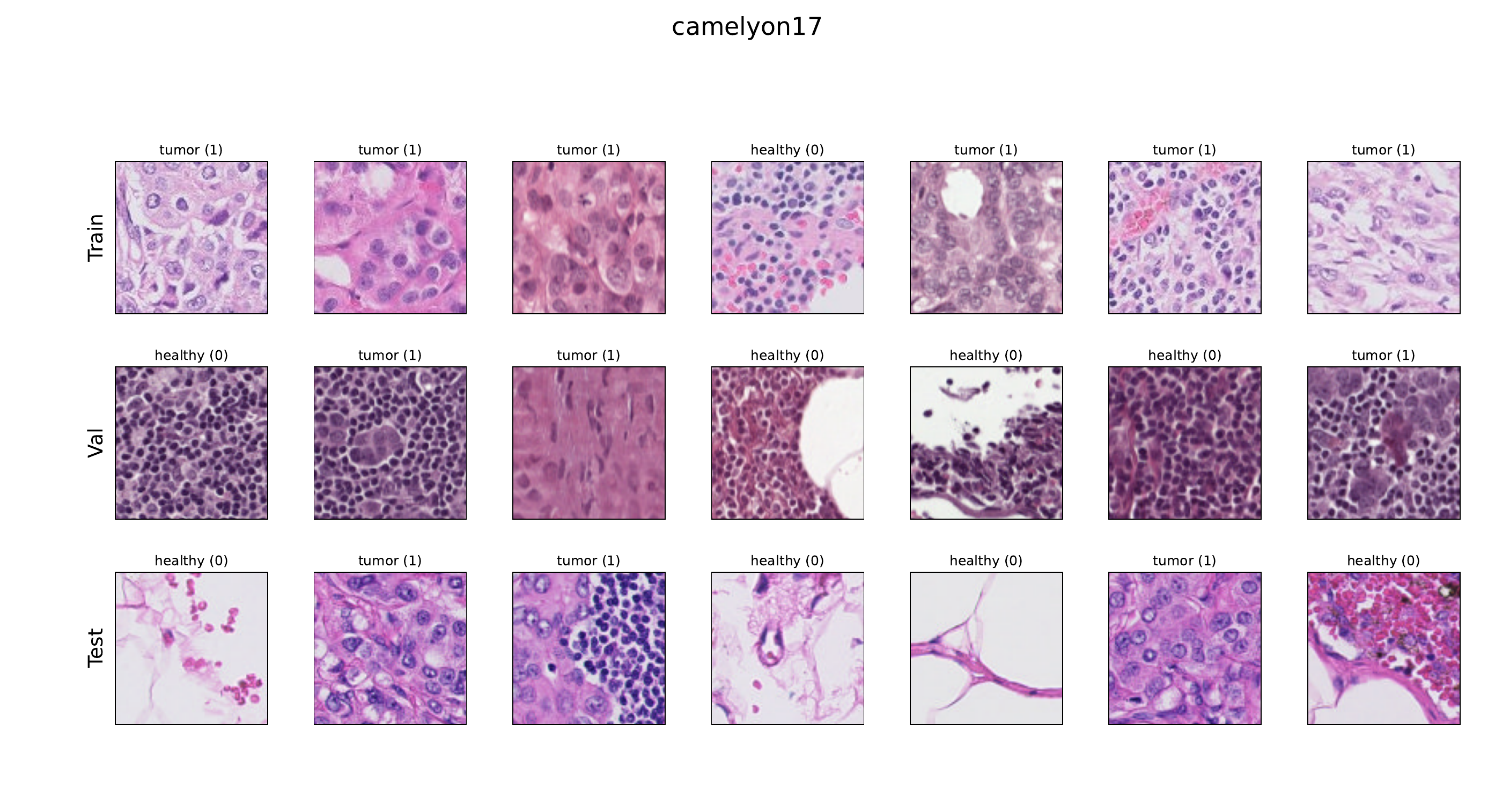}
        \caption{
        Examples from the camelyon17 dataset \cite{bandi2018detection}.
        The first row contains images from the training distribution, which consists of images from four hospital sites.
        The validation and test splits are composed of images from disjunct hospitals, therefore, introducing a domain shift.
        We additionally split the test data to obtain our tuning examples.
        }
        \label{fig:camelyon17-examples}
    \end{figure}

    Figure~\ref{fig:camelyon17-examples} visualizes images contained in the camelyon17 dataset \cite{bandi2018detection}.
    This dataset is composed of similar images from different hospital sites.
    These different capturing sites ensure a prominent domain shift between training and test distribution.
    To ensure our tuning examples contain information about the test distribution, we split the original test split provided by \cite{koh2021wilds} to generate our tuning batches.
    We provide more details about our specific setup in Section~\ref{sec:camelyon17}.

    \subsection{\uppercase{Baseline Overfitting}}
    \label{app:overfit}

    In addition to our change penalization, we propose a simple early-stopping scheme.
    We use this stopping criterion to reduce overfitting on the tuning data for our baseline methods.
    Here we now report what happens if we train the baseline approaches for $\epsilon = 50$ update steps longer.
    For this comparison, we use the dataset and setup described in Section~\ref{sec:mela}.

    \begin{figure}
        \centering
        \includegraphics[width=0.48\textwidth]{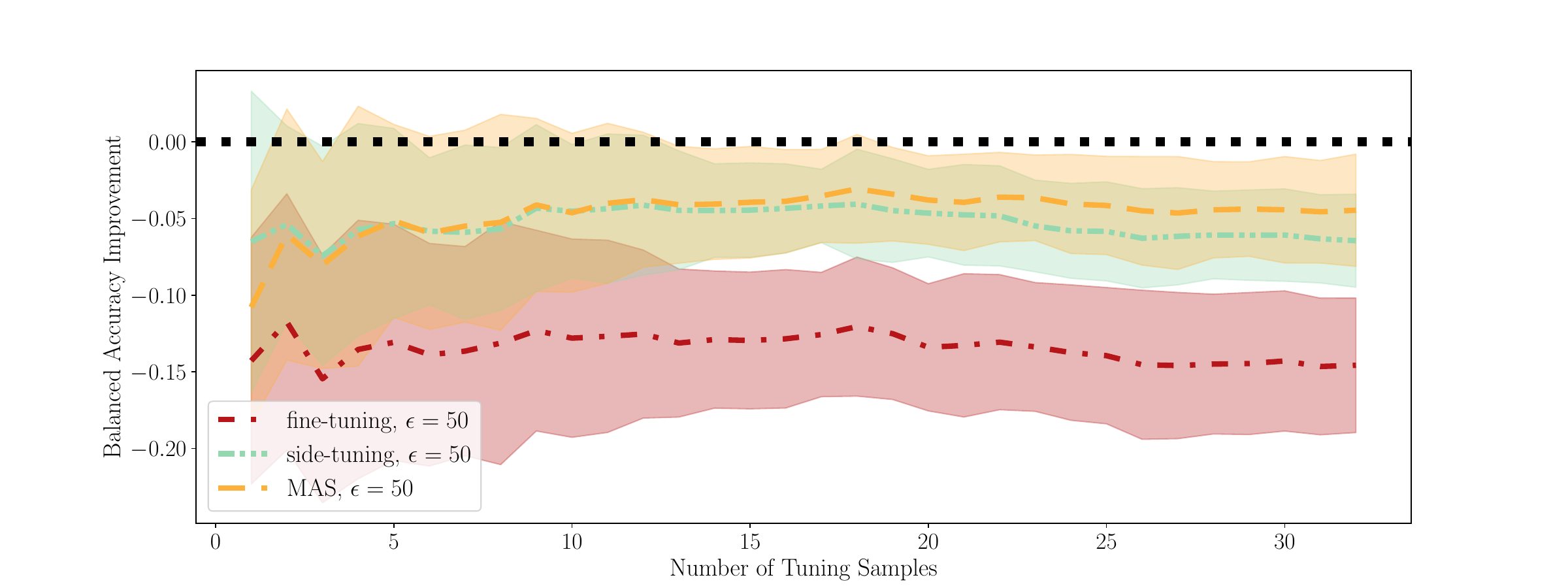}
        \caption{Baseline methods trained for 50 update steps longer than our stopping criteria indicates. The black dotted line indicates no change in balanced accuracy. All methods lead to a decrease in performance.}
        \label{fig:overfit}
    \end{figure}

    Figure~\ref{fig:overfit} shows the results for the amount of tuning samples between 1 and 32.
    We see that the performances of the pre-trained models on the test distribution deteriorate instead of increase.
    Hence, we conclude that training for too long leads to overfitting and reduces model performance significantly.
    However, we can see that both side-tuning and especially MAS, work better than classical fine-tuning.
    MAS already implicitly penalizes change to some parameters, and side-tuning adds parameters and leaves the original parameters untouched.

    \subsection{Hyperparameter Selection}
    \label{app:hp-selection}

    Given computational constraints, we select the hyperparameters for our penalization approach on the melanoma classification problem described in Section~\ref{sec:mela}.
    Here, we describe the results of our small ablation study and discuss the influence of the three hyperparameters for our method: penalization norm, penalization intensity $\lambda$, and stopping parameter $\epsilon$.
    First, we analyze the first two hyperparameters first.
    For this, we set $\epsilon$ to five following the observation made in Section~\ref{sec:impl-details}.
    We test our approach for the $\ell_1$, $\ell_2$, and $0.5 \cdot (\ell_1 + \ell_2)$ norms.
    Henceforth, we abbreviate the latter with $\ell_1 + \ell_2$.

    \begin{figure}
        \centering
        \begin{subfigure}{0.48\textwidth}
            \centering
            \includegraphics[width=\textwidth]{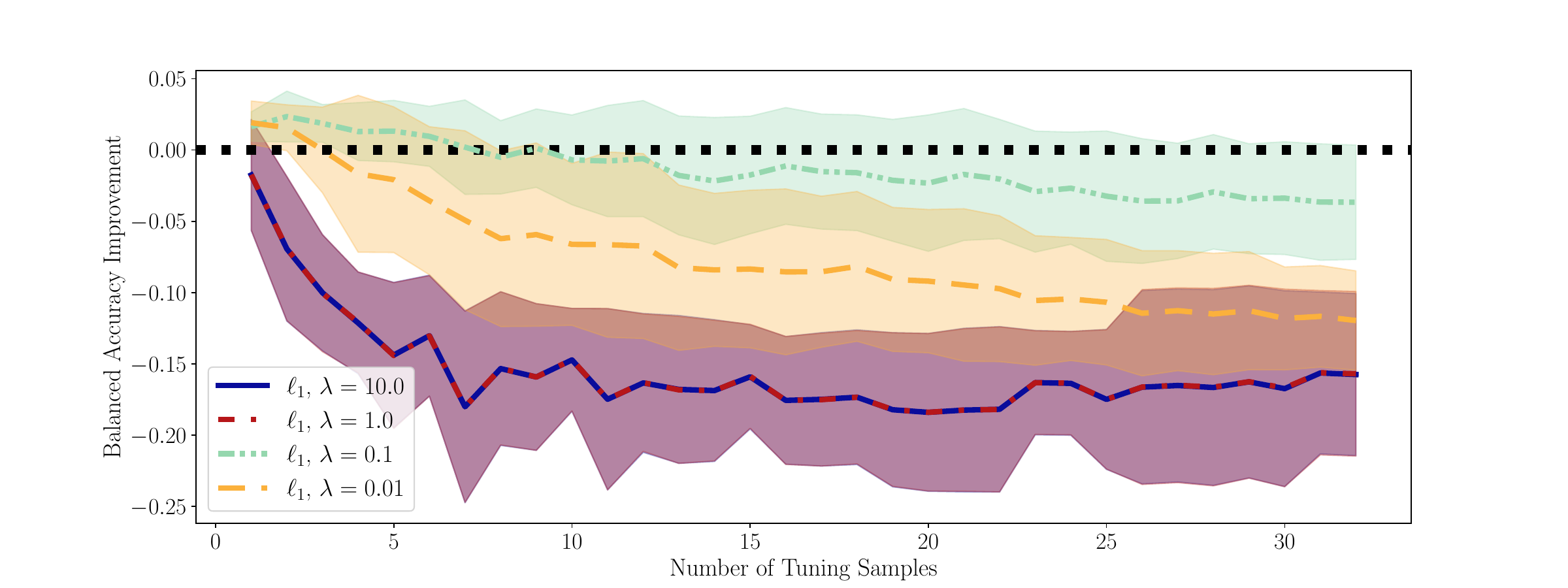}
            \caption{$\ell_1$ norm penalization.}
            \label{fig:l1}
        \end{subfigure}
        \begin{subfigure}{0.48\textwidth}
            \centering
            \includegraphics[width=\textwidth]{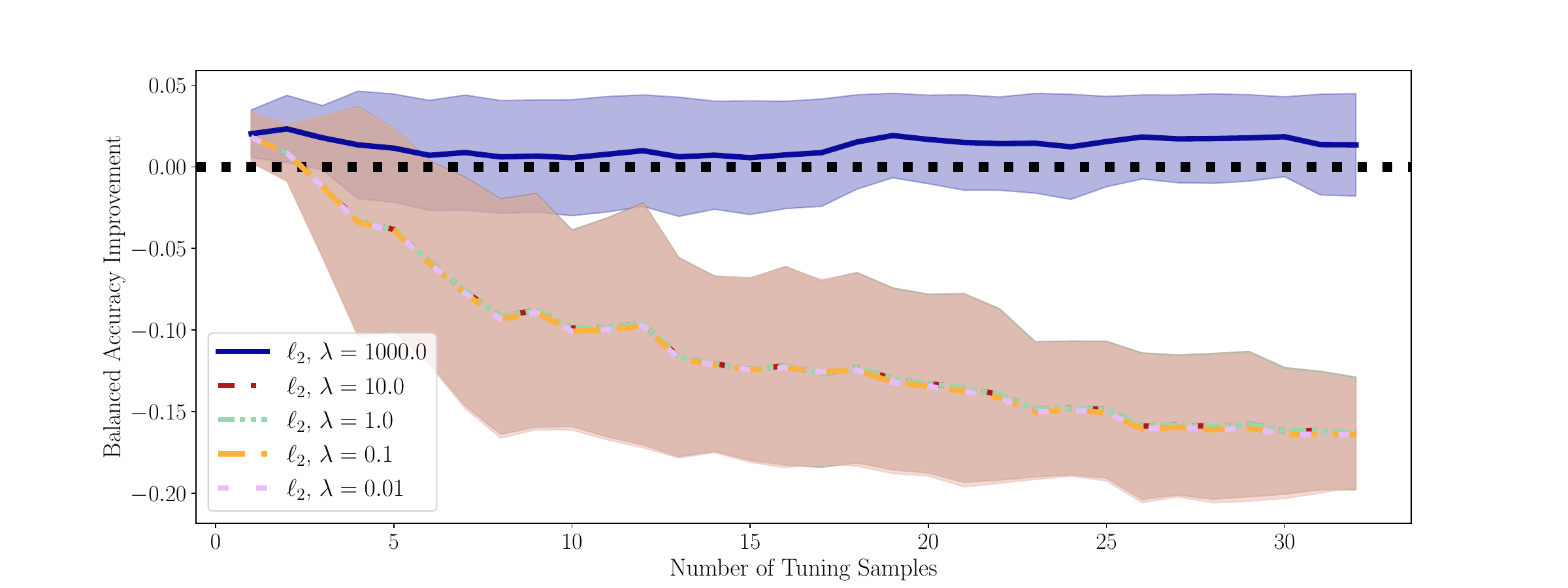}
            \caption{$\ell_2$ norm penalization.}
            \label{fig:l2}
        \end{subfigure}
        \begin{subfigure}{0.48\textwidth}
            \centering
            \includegraphics[width=\textwidth]{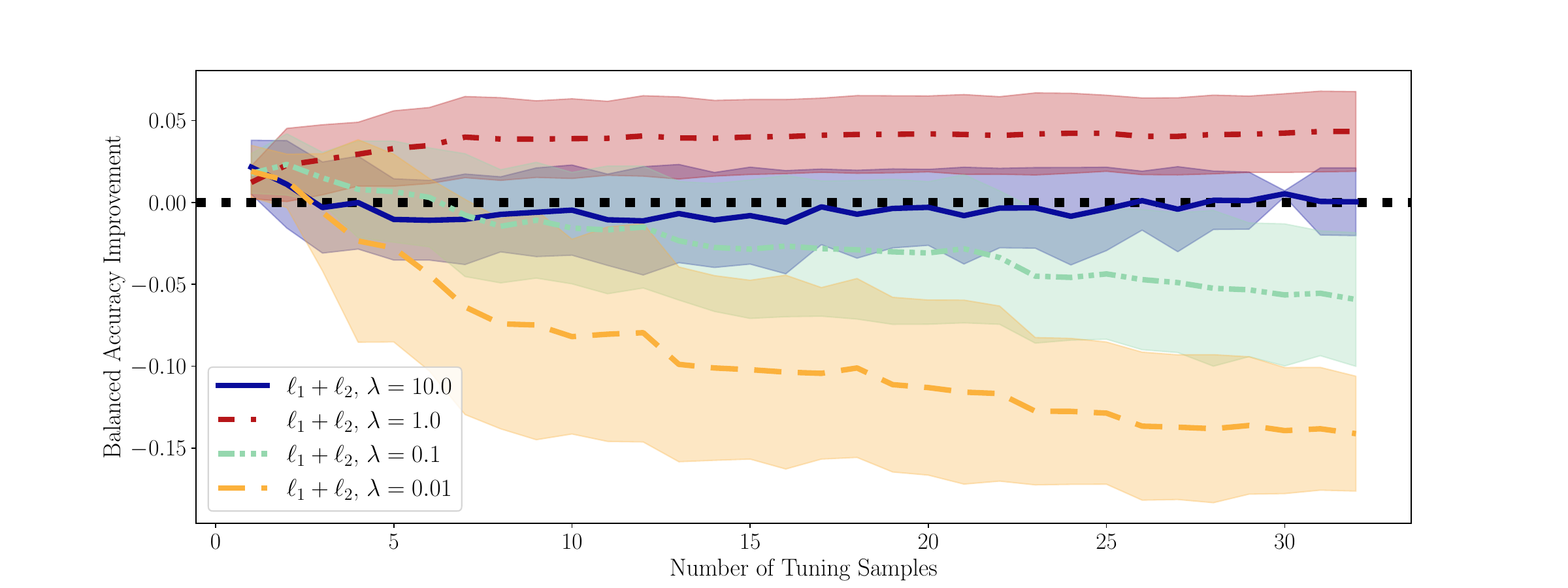}
            \caption{$\ell_1 + \ell_2$ norm penalization.}
            \label{fig:l1l2}
        \end{subfigure}
        \caption{
        Results for different parameter change norms. 
        For each norm, we visualize varying settings of the change penalization parameter $\lambda$.
        For this evaluation, we set $\epsilon = 5$ to ensure that the change penalization is larger than zero, following the observations made in Section~\ref{sec:impl-details}.
        }
        \label{fig:norm-lambda}
    \end{figure}

    Figure~\ref{fig:norm-lambda} shows varying settings of the parameter $\lambda$ for each of the three norms.
    First, we can see that for the $\ell_1$ norm and higher $\lambda$ values, the tuning deteriorates completely.
    Remember that $\ell_1$ regularization leads to sparse solutions \cite{goodfellow2016deep}, i.e., the optimum for many parameters is zero.
    Hence, only tuning a small number of parameters (for large settings of $\lambda$) seems to be counterproductive.
    In contrast, it works better for smaller $\lambda$ settings, especially for smaller numbers of tuning samples, e.g., $b=1$.
    Nevertheless, for larger amounts of tuning samples, performance still decreases.
    Using the $\ell_2$ norm also results in an increase in performance for smaller numbers of tuning samples given $\lambda$ settings of 10 and lower.
    In fact, all of these runs lead to very similar parameters and, therefore, performance differences.
    We argue that this behavior is overfitting similar to the overfitting of long tuning observed in Appendix~\ref{app:overfit} indicated by the similar performance drop.
    Hence, we additionally ran a set of experiments with $\lambda = 1000.0$ for $\ell_2$ norm.
    For this setting, the models, on average, increase in performance.
    However, this setting leads to convergence without correcting all tuning examples for large $b$, given the strong change penalization.
    We, therefore, stop the tuning after 500 update steps.
    Finally, Figure~\ref{fig:l1l2} visualizes the behavior of our change penalization approach for the combined $\ell_1$ and $\ell_2$ norms.
    Here we see that for very high penalization ($\lambda = 10$), the model performance for larger batch sizes stays nearly constant compared to the pre-trained model.
    Hence, penalization prevents change.
    Values for $\lambda$ lower than 1.0 deteriorate the model performance, especially for larger $b$.
    For $\lambda = 0.01$, our approach is closely related to standard fine-tuning.
    Here the setting $\epsilon = 5$ could already lead to overfitting on the tuning batches as observed in Appendix~\ref{app:overfit}.
    In contrast, $\lambda = 1.0$ leads to an improvement in the performance metric for all analyzed numbers of tuning samples.
    This effect is more pronounced for batch sizes 7 and up.

    To conclude: In this first hyperparameter ablation, we observe that in our setup for the melanoma classification task described in Section~\ref{sec:mela}, the $\ell_1 + \ell_2$ norm works best overall.
    Here we select $\lambda = 1.0$ as the penalization parameter.
    \begin{figure}
        \centering
        \includegraphics[width=0.48\textwidth]{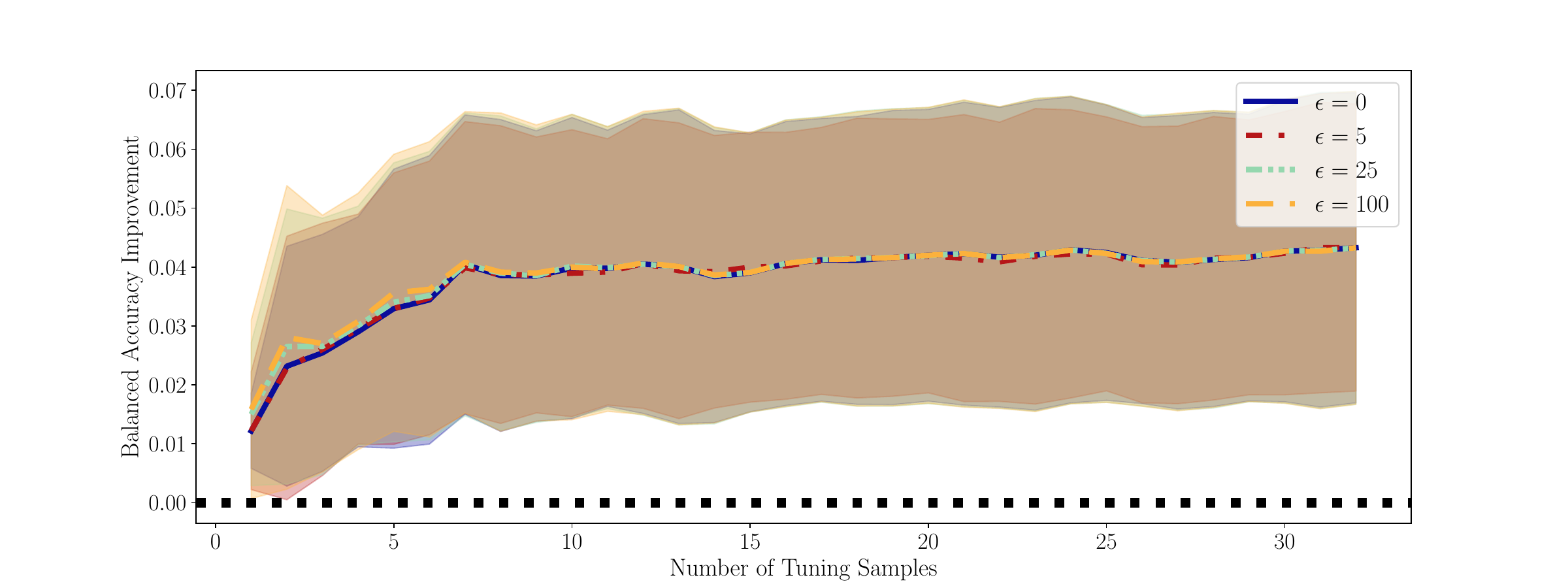}
        \caption{
        Different settings for the stopping parameter $\epsilon$. 
        Here we investigate our approach using the $\ell_1 + \ell_2$ norm together with $\lambda = 1.0$.
        }
        \label{fig:epsilon}
    \end{figure}
    For this setup, we now investigate different settings of the stopping criterion $\epsilon$.
    Figure~\ref{fig:epsilon} visualizes results for $\epsilon$ values between 0 and 100.
    Our chosen set of hyperparameters is robust for all settings of $\epsilon$.
    All runs converge to indistinguishable performance increases.
    Hence, we select $\epsilon = 0$ for our main experiments to increase comparability with the extended baseline methods.
    
\fi
\end{document}